\let\oldyear\year
\documentclass{ieeeaccess}
\usepackage{spotcolor}

\let\year\oldyear
\usepackage{cite}

\newcommand{\headerlogoall}{}  
\newcommand{\headerlogo}{}     
\usepackage{multirow}

\usepackage{adjustbox} 
\usepackage{subcaption}
\usepackage{amsmath,amssymb,amsfonts}
\usepackage{booktabs}
\usepackage{graphicx}
\usepackage{textcomp}
\def\BibTeX{{\rm B\kern-.05em{\sc i\kern-.025em b}\kern-.08em
    T\kern-.1667em\lower.7ex\hbox{E}\kern-.125emX}}

\usepackage{graphicx}
\usepackage{tikz}
\usetikzlibrary{quantikz2}

\graphicspath{ {./figures/} }
\usepackage{hyperref}
\usepackage{float}
\usepackage{verbatim} 
\usepackage{placeins}
\usepackage{savesym}

\usepackage{textcomp}

\usepackage{savesym}
\savesymbol{push}
\usepackage{array}
\usepackage{tikz,tikzscale}
\usepackage{pythonhighlight}

\usepackage{algorithm}
\usepackage{algpseudocode}
\usepackage{amsmath}

\usetikzlibrary{shapes,arrows,positioning}
\tikzset{
    startstop/.style={ellipse, draw, minimum width=3cm, minimum height=1cm, text centered, font=\normalsize},
    process/.style={rectangle, draw, minimum width=8cm, minimum height=1cm, text centered, font=\normalsize, text width=7.5cm},
    arrow/.style={thick,->,>=stealth}
}
\usepackage{tipa}        
\usepackage{microtype}
\usepackage{xcolor}

\usetikzlibrary{backgrounds}
\pgfkeys{/donut/.cd,
inner radius/.initial=0.7cm,
inner radius=0.7cm,
outer radius/.initial=3.14cm,
outer radius=3.14cm,
text color/.initial=white,
text color=white}
\newcommand{\donutchart}[2][]{
   \pgfmathsetmacro{\totalnum}{0}
   \foreach [count=\n] \value/\colour/\name in {#2} {
     \pgfmathparse{\value+\totalnum}
     \global\let\totalnum=\pgfmathresult
     \xdef\numitems{\n}
   }

  \begin{tikzpicture}
  \pgfmathsetmacro{\wheelwidth}{\pgfkeysvalueof{/donut/outer
  radius}-\pgfkeysvalueof{/donut/inner radius}}
  \pgfmathsetmacro{\midradius}{(\pgfkeysvalueof{/donut/outer radius}
  +\pgfkeysvalueof{/donut/inner radius})/2}

  \begin{scope}[#1]

    \pgfmathsetmacro{\cumnum}{0}
    \foreach \value/\colour/\name in {#2} {
        \pgfmathsetmacro{\newcumnum}{\cumnum + \value/\totalnum*360}

        \pgfmathsetmacro{\midangle}{-(\cumnum+\newcumnum)/2}
        \begin{scope}[on background layer]
          \filldraw[draw=white,fill=\colour]
          (-\cumnum:\pgfkeysvalueof{/donut/outer radius})
          arc(-\cumnum:-(\newcumnum):\pgfkeysvalueof{/donut/outer radius}) --
          (-\newcumnum:\pgfkeysvalueof{/donut/inner radius})
          arc(-\newcumnum:-(\cumnum):\pgfkeysvalueof{/donut/inner radius}) -- cycle;
        \end{scope}
        \draw node [text=\pgfkeysvalueof{/donut/text color},
        font=\bfseries\sffamily] at
        (\midangle:{\pgfkeysvalueof{/donut/inner radius}+\wheelwidth/2}) {\name};

        \global\let\cumnum=\newcumnum
    }

  \end{scope}

  \end{tikzpicture}}

\tikzset
{
    treenode/.style = {circle, draw=black, align=center, minimum size=1cm},
    subtree/.style  = {isosceles triangle, draw=black, align=center, minimum height=0.5cm, minimum width=1cm, shape border rotate=90, anchor=north}
}

\usetikzlibrary{positioning,quotes}

\NewSpotColorSpace{PANTONE} 

\AddSpotColor{PANTONE}{PANTONE3015C}{PANTONE\SpotSpace 3015\SpotSpace C}{0 0 0 1} 

\SetPageColorSpace{PANTONE}%

\definecolor{accessblue}{cmyk}{0,0,0,1}   
\definecolor{greycolor}{cmyk}{0,0,0,1}    

\begin{document}

\title{Automatic Speech Recognition for the Basa\`{a} Language: A Low-Resource Approach}
\author{\uppercase{Sophie Gertrude Ngo Mock}\authorrefmark{1},  \uppercase{Charles Moudina Varmantchaonala} \authorrefmark{2} \IEEEmembership{Member, IEEE}, \uppercase{Paul Dayang} \authorrefmark{1}, \uppercase{Jean Michel Nlong II}\authorrefmark{1} and \uppercase{Christopher Gies}\authorrefmark{2}}


\address[1]{Department of Mathematics and Computer Science, Faculty of Science, University of Ngaoundere, Ngaoundere, 454, Cameroon} 
\address[2]{Institute for Physics, Faculty V, Carl von Ossietzky University of Oldenburg, 26129 Oldenburg, Germany}



\corresp{Corresponding author: Paul Dayang (e-mail: pdayang@univ-ndere.cm)}

\begin{abstract}
    The rapid advancement of Artificial Intelligence (AI) and Natural Language Processing (NLP) has revolutionized the way humans interact with machines. Among the most impactful developments is Automatic Speech Recognition (ASR), which enables computers to convert spoken language into text. Systems such as those built on deep neural networks, transformer architectures, and self-supervised learning have achieved near-human performance for well-resourced languages such as English and French. Yet, these advances have disproportionately benefited a small fraction of the world's languages.
\end{abstract}


\titlepgskip=-21pt

\maketitle

\section{Introduction}
 
The rapid advancement of Artificial Intelligence (AI) and Natural Language Processing (NLP)
has revolutionized the way humans interact with machines. Among the most impactful
developments is Automatic Speech Recognition (ASR), which enables computers to convert
spoken language into text. Systems such as those built on deep neural networks, transformer
architectures, and self-supervised learning have achieved near-human performance for
well-resourced languages such as English and French~\cite{baevski2020wav2vec,vaswani2017attention}.
Yet, these advances have disproportionately benefited a small fraction of the world's
languages.
 
Africa is home to more than 2\,000 languages, many of which possess rich oral traditions
and serve as primary means of communication for millions of people~\cite{alabi2025charting} .
However, the vast majority of African languages, including Basa\`{a}, remain severely
under-resourced in the digital domain. As defined by~\cite{besacier2014asr}, an
under-resourced language is one that lacks adequate digital linguistic resources, including
normalized orthography, annotated corpora, phonetic dictionaries, and computational tools.
The Basa\`{a} language, a Bantu language spoken by approximately 300\,000 people in the
Centre and Littoral regions of Cameroon, exemplifies this digital marginalization. Despite
its cultural significance and growing urban speaker community, Basa\`{a} has no known prior
ASR system.
 
The absence of ASR tools for Basa\`{a} represents both a technological gap and a cultural
challenge. In a world where voice interfaces are increasingly embedded in everyday
life, from virtual assistants to healthcare applications and educational
platforms, communities that speak underrepresented languages are effectively excluded from
these technological benefits. Bridging this gap is not merely a matter of technical
interest; it is a question of linguistic equity and digital inclusion.
 
The central research question driving this work is: How can modern ASR techniques
be effectively applied to a low-resource language like Basa\`{a}, where annotated data is
scarce, orthographic standards are limited, and the language presents typologically complex
features such as tonal contrasts and a rich morphosyntactic system.
 
To address this question, this paper makes the following contributions:
 
\begin{itemize}
  \item We provide a comprehensive morphosyntactic and phonological analysis of the Basa\`{a}
        language, identifying features that are directly relevant to ASR system design,
        such as tone, vowel quantity, and the consonant inventory.
 
  \item We present the first end-to-end ASR pipeline for Basa\`{a}, based on fine-tuning the
        XLS-R model, a large-scale multilingual self-supervised speech model, on a
        Basa\`{a} -- French corpus from Mozilla Common Voice.
 
  \item We report quantitative evaluation results (WER\,=\,14.13\%, CER\,=\,3.51\%) on a
        test set of 1\,671 utterances, establishing a first baseline for ASR research on
        Basa\`{a}.
 
  \item We present a prototype mobile application (\textit{Mahop}) integrating the ASR
        model, enabling real-world speech transcription and crowdsourced data collection.
\end{itemize}
 
The remainder of this paper is organized as follows. Section~\ref{sec:linguistic} presents
a linguistic analysis of Basa\`{a}, focusing on features relevant to ASR. Section~\ref{sec:model}
describes the proposed ASR model architecture and its components. Section~\ref{sec:evaluation}
presents the evaluation methodology, experimental results, and a discussion of findings.
Section~\ref{sec:conclusion} concludes the paper with future directions.
 
\section{Analysis of the Basa\`{a} Language}
\label{sec:linguistic}
 
A thorough understanding of the linguistic structure of a target language is a prerequisite
for designing effective ASR systems. The phonological, morphological, and syntactic
properties of Basa\`{a} directly influence acoustic modeling decisions, tokenization
strategies, and the expected difficulty of recognition. In this section, we present the key
characteristics of the Basa\`{a} language relevant to speech processing.
 
\subsection{General Overview of Basa\`{a}}
 
\begin{figure}[!ht]
\centering
\includegraphics[width=0.7\linewidth]{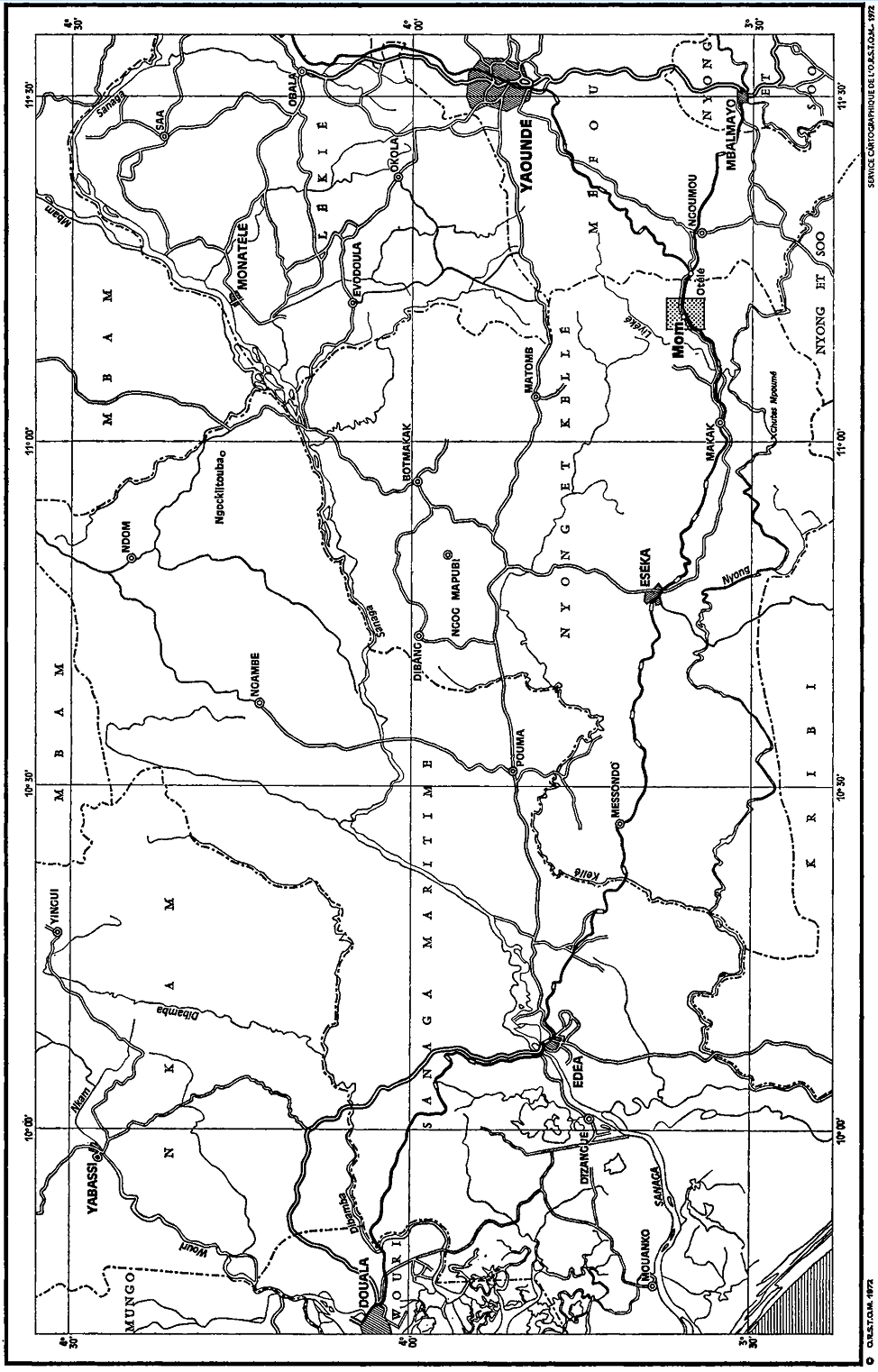}
\caption{An overview of the regions of Cameroon where the Basa\`{a} language is spoken, adapted from \cite{champaud2021mom}.}
\label{fig:african_phyla}
\end{figure}

Basa\`{a} (also spelled \textit{b\`{a}s\`{a}a} or Bassa) is a Bantu language belonging to
the Niger-Congo language family. Figure~\ref{fig:african_phyla} illustrates the regions of Cameroon where the Basaà language is spoken. It is classified under the Banto\"{i}de-Sud group and
bears the Guthrie code A.43a~\cite{makasso2008intonation}. The language is spoken
predominantly in two regions of Cameroon: the Centre Region (Nyong and Kell\'{e} department)
and the Littoral Region (Wouri, Sanaga Maritime, and Nkam departments). Urbanization has
also established significant Basa\`{a}-speaking communities in Douala and Yaound\'{e}, as
well as among the Cameroonian diaspora in Europe and North America.
 
Ethnologists \footnote{https://www.ethnologue.com/language/bas/\#typology} \cite{makasso2008intonation} identify approximately twelve dialectal varieties of Basa\`{a}, including
Bakem, Bibeng, Diboum, Log, Mpo, Mbang, Ndokama, Basso, Ndokbele, Ndokpenda, and
Nyamtam. This dialectal diversity presents a challenge for ASR systems, which must ideally
account for phonological and lexical variation across dialects. The present work focuses on
the Mpo dialect, specifically the Basa\`{a} Babimbi spoken in the Ngamb\`{e} region, which
constitutes the primary dialect represented in the Mozilla Common Voice corpus used for
this study.
 
Sociolinguistically, Basa\`{a} occupies an intermediate position: it functions as a
vehicular language in trade and transport across Cameroonian communities, yet it enjoys no
official status under Cameroon's constitution, which recognizes only French and English as
official languages~\cite{kamdem2025decolonizing}. Despite this institutional
marginalization, organizational initiatives such as CABTAL (Cameroon Association for Bible
Translation and Literacy) have contributed to orthographic standardization through biblical
translations, providing a limited but valuable textual resource base.

\subsection{Morphosyntactic Analysis of Basa\`{a}}

This part describes some key aspects of Basaà such as the phonological system, the morphological properties and the syntactic propertied.

\subsubsection{Phonological System}
 
The phonological system of Basa\`{a} presents several features that directly impact ASR
system design. Understanding these features is critical for making informed decisions about
the tokenization unit (phoneme-level vs.\ character-level), the design of the vocabulary,
and the interpretation of recognition errors.
 
Basa\`{a} possesses a seven-vowel inventory comprising three front vowels (/i/, /e/, /\textipa{E}/),
three back vowels (/u/, /o/, /\textipa{O}/), and one central low vowel (/a/)~\cite{makasso2008intonation,zribi1989preliminaires}. These vowels are fully represented in the symbol inventory of the General Alphabet of Cameroonian Languages (GACL), a reference orthographic system designed to harmonize the transcription of Cameroon's indigenous languages, from which Basa\`{a} draws its graphemes for its standardized writing system ~\cite{nzali2022basaa}.
A phonologically distinctive feature of the Basa\`{a} vowel system is the opposition
between short and long vowels, which carries lexical meaning. For example, the short vowel
/i/ in \textit{t\'{i}} (to give) contrasts with the long vowel /ii/ in \textit{t\'{i}\'{i}}
(to touch). This quantity distinction implies that ASR systems must be sensitive to vowel
duration, a feature that is often challenging to capture with standard acoustic models
trained on languages without such distinctions.
 
\begin{table}[htbp]
  \centering
  \caption{Short/long vowel oppositions in Basa\`{a}~\cite{makasso2008intonation}}
  \label{tab:vowels}
  \renewcommand{\arraystretch}{1.2}
  \begin{tabular}{clllll}
    \toprule
    & \textbf{Vowel} & \textbf{Short Form} & \textbf{Meaning}
      & \textbf{Long Form} & \textbf{Meaning} \\
    \midrule
    I & /i/ & t\'{i}      & to give  & t\'{i}\'{i}   & to touch  \\
    U & /u/ & h\'{u}      & audacity & h\'{u}\'{u}   & to return \\
    E & /e/ & \'{e}       & to clear land & p\'{e}\'{e} & viper  \\
    O & /o/ & s\'{o}      & antelope & s\`{o}\`{o}   & to hide   \\
    \textipa{E} & /\textipa{E}/ & \'{e} & tree & \textipa{E}\textipa{E} & to cry \\
    A & /a/ & b\`{a}      & to skin  & b\`{a}\`{a}   & to filter \\
    \bottomrule
  \end{tabular}
\end{table}
 
The consonant inventory of Basa\`{a} is rich and typologically marked. According
to~\cite{makasso2015basaa}, it comprises 30 consonants including plosives, nasals,
fricatives, affricates, implosives (notably /\textipa{b}/), and pre-nasalized consonants.
The presence of implosives (/\textipa{b}/) and pre-nasalized stops
(/\textsuperscript{m}b/, /\textsuperscript{n}d/, /\textsuperscript{\textipa{N}}g/)
represents a significant challenge for ASR systems trained primarily on European
languages, as these sounds may not have close acoustic equivalents in the pre-training
data. The implication for ASR design is the need for a character-level or phoneme-level
vocabulary that explicitly encodes these distinctive sounds.
 
The syllabic structure of Basa\`{a} includes the following patterns: V, VV, CV, VC, CVV,
and CVC~\cite{makasso2008intonation}. This variety of syllable types, including
vowel-initial syllables and closed syllables with long vowels, creates phonotactic
constraints that an ASR system must implicitly learn to model.
 
Crucially, Basa\`{a} is a tonal language with two simple tones: a high
tone (marked with an acute accent: [$\acute{\,}$]) and a low tone
(marked with a grave accent: [$\grave{\,}$]). In certain phonological
contexts, the low tone can manifest as a rising contour, and the high
tone as a falling or mid-level tone~\cite{makasso2008intonation}. Tone
in Basa\`{a} is lexically and grammatically distinctive: two words
differing only in their tonal pattern can have entirely different
meanings. This tonal dimension adds a layer of complexity to ASR, as
spectral features alone may be insufficient to distinguish tonal
minimal pairs. Integrating tonal information into the ASR pipeline
either through specialized acoustic features or post-processing with
a tonal language model is therefore an important avenue for future work.
 
\subsubsection{Morphological Properties}
 
Basa\`{a} is an agglutinative language with a well-developed system of nominal classes, a
characteristic feature of Bantu languages. Nouns are organized into singular-plural pairs
determined by noun class membership, and this classification governs agreement patterns
throughout the sentence. Unlike the masculine/feminine gender distinction in French, Bantu
noun classes are semantically and formally diverse, and the same lexical root can appear
with different morphosyntactic properties depending on its class prefix.
 
The verb system in Basa\`{a} is particularly rich. According to Bitja'a Kody (1990) as
cited in~\cite{makasso2008intonation}, the morphological structure of a verbal theme
follows the pattern: \textit{Verbal Radical + Extensions + Final Vowel}. Verbal extensions
encode a wide range of grammatical meanings including applicative, passive, reflexive,
frequentative, direct causative, indirect causative, simultaneous, associative, and
possessive forms. This means that from a single root verb such as \textit{t\`{e}\textipa{N}}
(to attach), a large number of morphologically derived forms can be produced, each with a
distinct pronunciation. For ASR, this productivity implies a large and open vocabulary,
making word-level language models less practical and strengthening the case for
character-level modeling with connectionist temporal classification (CTC) decoding.
 
Furthermore, vowel alternations occur in verbal radicals under the influence of extensions:
mid vowels /e/ and /o/ raise to /i/ and /u/ respectively, and /\textipa{E}/ and /a/ raise
to /e/, while /\textipa{O}/ raises to /o/. This regular but non-trivial morphophonological
process affects the acoustic realization of verb roots and must be handled by the ASR
acoustic model.
 
\subsubsection{Syntactic Properties}
 
The basic word order in Basa\`{a} is Subject-Verb-Object (SVO), with the subject pronoun
always preceding the verb. Demonstratives and possessive suffixes agree with the noun class
of the head noun. The language employs two coordinative conjunctions, [\textit{n\`{\i}}]
and [\textit{n\`{a}}], both translatable as `and/with', with [\textit{n\`{\i}}] used for
independent association and [\textit{n\`{a}}] for joint action or inclusive reference.
 
Personal pronouns in Basa\`{a} vary according to their grammatical function (subject,
object, emphatic) and according to the noun class of the referent. Following the first two
persons (singular and plural), all remaining classes correspond to the third person with
singular and plural forms for each class. This complex pronominal system, along with the
rich verbal morphology, implies that ASR systems for Basa\`{a} must handle a wide range of
surface forms for related semantic content, a challenge that character-level CTC modeling
is well-suited to address.
 
\section{Proposed ASR Model for Basa\`{a}}
\label{sec:model}
 
This section presents the architecture of our proposed ASR system for Basa\`{a}. We
describe the overall pipeline from raw audio input to text output, provide an overview of
the model architecture, and detail each component and its role, highlighting the specific
adaptations made to address the linguistic characteristics of Basa\`{a} identified in
Section~\ref{sec:linguistic}.
 
\subsection{Overview of the ASR Pipeline}
 
Our ASR system follows an end-to-end architecture based on fine-tuning the XLS-R
model~\cite{babu2021xlsr}, a cross-lingual speech representation model derived from
Wav2Vec~2.0~\cite{baevski2020wav2vec}. The pipeline encompasses four major stages:
(1)~data collection and preprocessing, (2)~model configuration and adaptation,
(3)~CTC-based training, and (4)~inference and evaluation.
 
The overall architecture of the model, in Figure \ref{fig:pipeline}, can be summarized as follows: a raw audio signal in WAV format
(sampled at 16 \, kHz) is sent as input to a feature extractor (a convolutional neural
network) that produces a sequence of continuous latent representations. These representations
are then processed by a multi-layer transformer encoder, which captures long-range temporal
dependencies and cross-lingual acoustic patterns learned during pre-training. A linear
projection layer (the CTC head) maps the transformer outputs to a probability distribution
over the Basa\`{a} character vocabulary at each time step. During training, the CTC loss
function enables the model to learn the alignment between acoustic frames and character
sequences without requiring frame-level annotations. During inference, a greedy or beam
search decoder produces the final transcription.
 
The key insight motivating the use of XLS-R is that it was pre-trained on 436\,000 hours
of unlabeled speech data from 128 languages using self-supervised contrastive
learning~\cite{babu2021xlsr}. Although Basa\`{a} is not among the languages explicitly documented in the pre-training data of XLS-R, the model was trained on several Bantu languages, including Swahili, Kinyarwanda, Shona, and Zulu, among others ~\cite{babu2021xlsr}. These languages belong to the same linguistic family as Basa\`{a} and share several important phonological and morphological characteristics with it. This massive multilingual pre-training endows the model with rich, cross-lingual phonetic representations that generalize to new languages, including
low-resource ones, through a relatively small amount of fine-tuning data. This is
particularly relevant for Basa\`{a}, where labeled data is scarce. During the pre-training phase of the XLS-R model, it is important to note that the language's alphabet plays no role, as the model learns directly from raw acoustic speech signals without using any alphabet or textual transcriptions. The alphabet is only introduced during the fine-tuning stage, where the learned acoustic representations are mapped to the orthographic units of the target language ~\cite{baevski2020wav2vec,conneau2020unsupervised}.This shows that learning takes place before any transcription is introduced.
 
\subsection{Model Architecture and Components}
 In this section, we present the components of the proposed model, shown in Figure~\ref{fig:pipeline}, along with their respective functions.
\subsubsection{Input: Raw Audio Signal}
 
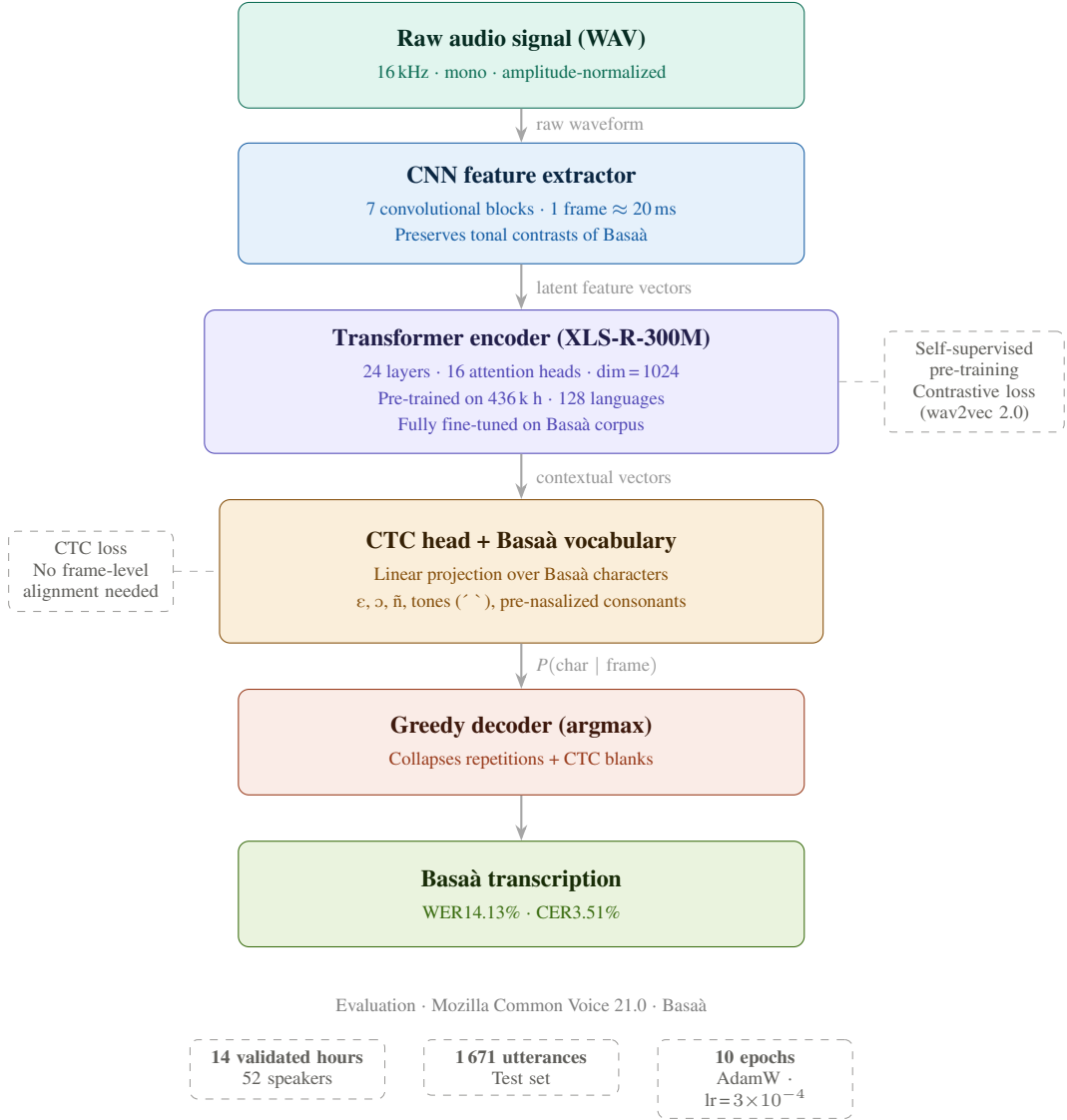
\begin{figure*}[!ht]
\centering
\definecolor{col1bg}{RGB}{225,245,238}  
\definecolor{col1bd}{RGB}{15,110,86}
\definecolor{col2bg}{RGB}{230,241,251}  
\definecolor{col2bd}{RGB}{24,95,165}
\definecolor{col3bg}{RGB}{238,237,254}  
\definecolor{col3bd}{RGB}{83,74,183}
\definecolor{col4bg}{RGB}{250,238,218}  
\definecolor{col4bd}{RGB}{133,79,11}
\definecolor{col5bg}{RGB}{250,236,231}  
\definecolor{col5bd}{RGB}{153,60,29}
\definecolor{col6bg}{RGB}{234,243,222}  
\definecolor{col6bd}{RGB}{59,109,17}
\definecolor{annotcol}{RGB}{95,94,90}   
 
\tikzset{
  stage/.style={
    draw, rounded corners=4pt, text centered,
    minimum width=7.5cm, font=\small,
    inner sep=6pt, align=center
  },
  annot/.style={
    draw=annotcol!60, dashed, rounded corners=3pt,
    font=\scriptsize, align=center, text=annotcol,
    inner sep=4pt
  },
  arr/.style={-{Stealth[length=6pt]}, thick, gray!70},
  darr/.style={dashed, gray!60, thin},
  lbl/.style={font=\scriptsize, text=gray!80, midway, right=2pt}
}
 
\begin{tikzpicture}[node distance=0.55cm]
 
\node[stage, fill=col1bg, draw=col1bd, text=col1bd!40!black,
      minimum height=1.4cm] (s1)
  {\textbf{Raw audio signal (WAV)}\\[2pt]
   \textcolor{col1bd}{\scriptsize 16\,kHz $\cdot$ mono $\cdot$ amplitude-normalized}};
 
\node[below=0.4cm of s1] (a1) {};
\draw[arr] (s1.south) -- node[lbl]{raw waveform} (s1.south |- a1.north)
           -- ++(0,-0.05cm);
 
\node[stage, fill=col2bg, draw=col2bd, text=col2bd!40!black,
      minimum height=1.6cm, below=0.45cm of s1] (s2)
  {\textbf{CNN feature extractor}\\[2pt]
   \textcolor{col2bd}{\scriptsize 7 convolutional blocks $\cdot$ 1 frame $\approx$ 20\,ms}\\
   \textcolor{col2bd}{\scriptsize Preserves tonal contrasts of Basa\`{a}}};
 
\draw[arr] (s2.south) -- node[lbl]{latent feature vectors} ++(0,-0.6cm);
 
\node[stage, fill=col3bg, draw=col3bd, text=col3bd!40!black,
      minimum width=8.4cm, minimum height=1.9cm,
      below=0.6cm of s2] (s3)
  {\textbf{Transformer encoder (XLS-R-300M)}\\[2pt]
   \textcolor{col3bd}{\scriptsize 24 layers $\cdot$ 16 attention heads $\cdot$ dim\,=\,1024}\\
   \textcolor{col3bd}{\scriptsize Pre-trained on 436\,k h $\cdot$ 128 languages}\\
   \textcolor{col3bd}{\scriptsize Fully fine-tuned on Basa\`{a} corpus}};
 
\node[annot, right=0.6cm of s3, text width=2.1cm] (ann3)
  {Self-supervised\\pre-training\\Contrastive loss\\(wav2vec 2.0)};
\draw[darr] (s3.east) -- (ann3.west);
 
\draw[arr] (s3.south) -- node[lbl]{contextual vectors} ++(0,-0.6cm);
 
\node[stage, fill=col4bg, draw=col4bd, text=col4bd!40!black,
      minimum width=8.0cm, minimum height=1.9cm,
      below=0.6cm of s3] (s4)
  {\textbf{CTC head + Basa\`{a} vocabulary}\\[2pt]
   \textcolor{col4bd}{\scriptsize Linear projection over Basa\`{a} characters}\\
   \textcolor{col4bd}{\scriptsize \textipa{E}, \textipa{O}, \~{n}, tones (\'{\,} \`{\,}),
     pre-nasalized consonants}};
 
\node[annot, left=0.6cm of s4, text width=1.9cm] (ann4)
  {CTC loss\\No frame-level\\alignment needed};
\draw[darr] (ann4.east) -- (s4.west);
 
\draw[arr] (s4.south) -- node[lbl]{$P(\text{char}\mid\text{frame})$} ++(0,-0.6cm);
 
\node[stage, fill=col5bg, draw=col5bd, text=col5bd!40!black,
      minimum height=1.4cm,
      below=0.6cm of s4] (s5)
  {\textbf{Greedy decoder (argmax)}\\[2pt]
   \textcolor{col5bd}{\scriptsize Collapses repetitions + CTC blanks}};
 
\draw[arr] (s5.south) -- ++(0,-0.6cm);
 
\node[stage, fill=col6bg, draw=col6bd, text=col6bd!40!black,
      minimum height=1.4cm,
      below=0.6cm of s5] (s6)
  {\textbf{Basa\`{a} transcription}\\[2pt]
   \textcolor{col6bd}{\scriptsize WER\;14.13\% $\cdot$ CER\;3.51\%}};
 
\node[below=0.55cm of s6, font=\scriptsize, text=gray] (footer)
  {Evaluation $\cdot$ Mozilla Common Voice 21.0 $\cdot$ Basa\`{a}};
 
\node[annot, below=0.25cm of footer, xshift=-3.1cm, text width=2.3cm] (m1)
  {\textbf{\small 14 validated hours}\\52 speakers};
\node[annot, below=0.25cm of footer, text width=2.3cm] (m2)
  {\textbf{\small 1\,671 utterances}\\Test set};
\node[annot, below=0.25cm of footer, xshift=3.1cm, text width=2.3cm] (m3)
  {\textbf{\small 10 epochs}\\AdamW $\cdot$ lr\,=\,$3\!\times\!10^{-4}$};
 
\end{tikzpicture}
\caption{End-to-end ASR pipeline for the Basa\`{a} language --- Fine-tuned XLS-R-300M.}
\label{fig:pipeline}
\end{figure*}

In Figure \ref{fig:pipeline}, the input to the system is a raw mono-channel audio file in WAV format, resampled to
16\,kHz. Audio preprocessing involves normalization of amplitude (to ensure consistent
input volume across speakers and recording conditions), conversion from MP3 to WAV
(required by the Hugging Face feature extractor), and padding or truncation to accommodate
variable-length utterances within a batch.
 
A key architectural decision is the use of raw waveforms as input, bypassing traditional
handcrafted features such as Mel-frequency cepstral coefficients (MFCCs). This choice is
justified by the findings of~\cite{baevski2020wav2vec}, who demonstrated that end-to-end
models operating on raw audio outperform MFCC-based systems, particularly in low-resource
settings. For a tonal language like Basa\`{a}, where fine-grained pitch information is
linguistically relevant, operating on raw waveforms preserves spectro-temporal details that
MFCC compression might discard.
 
\subsubsection{Feature Extraction: Multi-Layer CNN}
 
The first processing stage of XLS-R consists of a multi-layer convolutional neural network
(CNN) that maps the raw waveform into a sequence of latent feature vectors. These vectors
capture local acoustic patterns at a rate of approximately one frame per 20 milliseconds.
The CNN acts as a learned feature extractor, replacing traditional signal processing steps.
 
This stage processes the audio input before it reaches the transformer encoder. The output
of the CNN feature extractor is a sequence of dense vectors that encode the acoustic
content of the speech at a compressed temporal resolution. The specific architecture of the
convolutional feature extractor (number of layers, kernel sizes, strides) follows the
original Wav2Vec~2.0 specification as implemented in the Facebook AI Research (FAIR)
XLS-R-300M checkpoint.
 
\subsubsection{Transformer Encoder: Cross-Lingual Representation}
 
The core of the XLS-R model is a multi-layer transformer encoder with a multi-head
self-attention mechanism. The model variant used in this work (XLS-R-300M) contains
approximately 300 million parameters organized in 24 transformer layers, each with 16
attention heads and a hidden dimension of 1\,024. This large capacity enables the model to
capture complex, long-range acoustic dependencies.
 
Crucially, the pre-training of XLS-R used a contrastive self-supervised objective on audio
from 128 languages. During pre-training, the model learned to predict the correct quantized
speech unit from a set of distractors, given a masked context. This objective encourages
the model to develop phonetically meaningful, language-universal representations. For
Basa\`{a} specifically, although the language may not have been included in the
pre-training data, the linguistic proximity of Basa\`{a} to other Bantu languages, several
of which were represented in the XLS-R training set, provides a favorable starting point
for transfer.
 
Our contribution at this stage consists of fine-tuning all transformer layers on the
Basa\`{a} corpus. We observed that full fine-tuning (as opposed to freezing lower layers)
yielded better results, consistent with findings in the low-resource ASR
literature~\cite{conneau2020unsupervised}. This is likely because the acoustic properties
of Basa\`{a}, particularly its tonal contrasts and pre-nasalized consonants, require
adaptation of all representational levels.
 
\subsubsection{CTC Head and Basa\`{a} Vocabulary}
 
A linear projection layer (the CTC head) is appended to the transformer encoder. This
layer maps the transformer output at each time step to a probability distribution over the
Basa\`{a} character vocabulary. The vocabulary was constructed from the training corpus and
includes all unique characters appearing in Basa\`{a} transcriptions, including
diacritic-marked characters for tones (acute and grave accents), the \~{n} character, and
special characters specific to Basa\`{a} orthography.
 
Our contribution at this stage is the design of the Basa\`{a}-specific vocabulary and the
corresponding CTC tokenizer. The Basa\`{a} orthography includes tonal diacritics and
characters (such as \~{n}, \textipa{E}, \textipa{O})~\cite{makasso2008intonation,zribi1989preliminaires}
that must be explicitly represented. The tokenizer maps each character to an integer index,
and the CTC loss function enables the model to learn character-level alignments without
requiring explicit segmentation.
 
The CTC decoding strategy used during inference is greedy decoding (argmax at each time
step), which selects the most probable character at each frame and collapses repeated
characters and blank tokens. This simple approach proves effective given the character-level
granularity of the vocabulary.
 
\subsubsection{Architecture Specificity for Basa\`{a}: Comparison with French and English}
 
A central architectural motivation for this work is the observation that standard ASR
systems designed for French or English cannot be directly applied to Basa\`{a} without
significant adaptation. French and English benefit from massive labeled corpora (thousands
of hours), standardized orthographies, and absence of phonemic tone. Basa\`{a}, by
contrast, presents tonal distinctions, a small but growing corpus (approximately 14 hours
in Common Voice~21.0), pre-nasalized and implosive consonants, and high morphological
productivity.
 
The XLS-R architecture is particularly well-suited to bridge this gap because:
(1)~its self-supervised pre-training on 128 languages means it has already encountered
diverse phonetic inventories including some African languages;
(2)~its end-to-end character-level CTC formulation avoids the need for a lexicon or
pronunciation dictionary, which do not yet exist for Basa\`{a} in machine-readable form;
and (3)~the transfer learning approach allows the model to leverage its rich cross-lingual
representations while adapting to the specific acoustic and phonological properties of
Basa\`{a} through fine-tuning.
 
 
\section{Evaluation and Discussion}
\label{sec:evaluation}

This section focuses on the presentation of the Dataset, the Experimental Setup and hyper parameters, and the key Results we obtained.
 
\subsection{Dataset}
 
The corpus used in this work was collected from Mozilla Common Voice (version~21.0,
released March~2025), an open-source platform for multilingual speech data collection. The
Basa\`{a} dataset\footnote{https://mozilladatacollective.com/datasets/cmqigrvys00i3nr07ngkpb6b2} comprises approximately 241.59\,MB of MP3 audio recordings, totaling
14 recorded hours with 13 validated hours, collected from 52 speakers. Alongside the audio
files, the corpus includes a tab-separated value (TSV) file (\texttt{validated.tsv})
containing client identifiers, audio file paths, transcription sentences, up-vote and
down-vote counts, and speaker metadata.
 
The dataset was partitioned into three subsets: a training set (70\% of validated
utterances), a validation set (15\%), and a test set (15\%). The test set used for final
evaluation contains 1\,671 audio-transcription pairs. All audio files were resampled to
16\,kHz and normalized in amplitude before being passed to the model.
 
\subsection{Experimental Setup and Hyperparameters}
 
All experiments were conducted using GPU NVIDIA Tesla T4/P100, 25\,GB
RAM. The implementation used PyTorch and the Hugging Face Transformers and Datasets libraries.
 
Fine-tuning was performed for 10 epochs with the following hyperparameters: effective batch
size of 16 (batch size of 4 with gradient accumulation over 4 steps), learning rate of
$3\times10^{-4}$ with a linear warmup over 500 steps, AdamW optimizer, and CTC loss
function. These choices were guided by established best practices for low-resource ASR
fine-tuning \cite{conneau2020unsupervised,zhao2022improving}.
 
\begin{table}[htbp]
  \centering
  \caption{Evolution of WER on the validation set during training}
  \label{tab:training}
  \renewcommand{\arraystretch}{1.2}
  \begin{tabular}{rrrr}
    \toprule
    \textbf{Epoch} & \textbf{Train Loss} & \textbf{Val. Loss} & \textbf{WER (Val.)} \\
    \midrule
    500  & 1.3005 & 0.7516 & 93.77\% \\
    1000 & 0.5123 & 0.2769 & 60.90\% \\
    1500 & 0.3527 & 0.2222 & 52.34\% \\
    2000 & 0.2750 & 0.2069 & 50.12\% \\
    2500 & 0.2272 & 0.1727 & 43.29\% \\
    3000 & 0.1910 & 0.1617 & 40.54\% \\
    3500 & 0.1624 & 0.1414 & 37.43\% \\
    4000 & 0.1412 & 0.1422 & 36.11\% \\
    4500 & 0.1388 & 0.1342 & 33.71\% \\
    5000 & 0.1180 & 0.1298 & \textbf{14.13\%} \\
    \bottomrule
  \end{tabular}
\end{table}
 
\subsection{Results}
 
The final evaluation on the held-out test set of 1\,671 utterances yielded the following
performance metrics: Word Error Rate (WER)of 14.13 and Character Error Rate (CER) of 3.51. The WER of 14.13\% indicates that, on average, approximately 86 out of every 100 words are
correctly transcribed by the system. The substantially lower CER of 3.51\% reveals that
errors are predominantly localized to specific characters within words, rather than
wholesale word misrecognition. This gap between WER and CER is characteristic of
morphologically rich languages where small character-level errors, such as missing
diacritics or tone marks, can render an entire word incorrect at the word level while
leaving most characters intact.
 
\begin{table*}[htbp]
  \centering
  \caption{Sample transcriptions from the test set}
  \label{tab:samples}
  \renewcommand{\arraystretch}{1.2}
  \small
  \begin{tabular}{@{}c c c p{3.6cm} p{3.6cm} p{3.8cm}@{}}
    \toprule
     & \textbf{Time \ (sec.)} & \textbf{WER}
      & \textbf{Reference} & \textbf{Predicted} & \textbf{Error type} \\
    \midrule
    1 & 3.9 & 0.200
      & mudaa a \'{e}k ndap libii
      & mudaa a y\'{e}k ndap libii
      & Char.\ substitution (\'{e} $\to$ y\'{e}) \\
    2 & 4.0 & 0.375
      & u ntop b\'{e} nok ma\'{e}ba
      & u ntop b\'{e}nok may\'{e}ba
      & Missing space + vowel change \\
    3 & 3.9 & 0.000
      & men me nlok lipondo li mal\'{e}p
      & men me nlok lipondo li mal\'{e}p
      & Perfect transcription \\
    4 & 3.3 & 0.000
      & me nnek b\'{e} nye
      & me nnek b\'{e} nye
      & Perfect transcription \\
    5 & 3.8 & 0.500
      & bajo sa\~{n} ba mb\'{e}bna
      & bajosa\~{n} ba mb\'{e}bna
      & Missing word boundary \\
    6 & 3.0 & 0.000
      & nya\~{n} a y\'{e} mal\'{e}t
      & nya\~{n} a y\'{e} mal\'{e}t
      & Perfect transcription \\
    7 & 2.6 & 0.000
      & yom yem i
      & yom yem i
      & Perfect transcription \\
    8 & 3.1 & 0.250
      & a nsihil mut isi
      & a nsihil mut hisi
      & Consonant insertion (h) \\
    \bottomrule
  \end{tabular}
\end{table*}
 
The qualitative analysis in Table~\ref{tab:samples} reveals several recurring patterns of
errors. First, word boundary errors (missing spaces between tokens) account for a
meaningful proportion of word-level mistakes; these are likely related to the agglutinative
morphology of Basa\`{a}, where prefixes and roots are frequently written together in
natural usage. Second, diacritic-related substitutions (e.g., \'{e} $\to$ y\'{e}) suggest
that tonal and phonemic distinctions encoded in diacritics remain challenging for the model,
consistent with the known difficulty of tonal language ASR. Third, occasional consonant
insertions (e.g., \textit{isi} $\to$ \textit{hisi}) point to acoustic ambiguity between
certain segments. Importantly, a significant proportion of test utterances receive perfect
transcriptions (WER\,=\,0.0), demonstrating that the model has successfully generalized to
recognizable Basa\`{a} speech patterns.
 
\subsection{Discussion}
 
The results reported here must be interpreted in light of the specific challenges of the
Basa\`{a} ASR task and the broader low-resource ASR literature.
 
Regarding performance, a WER of 14.13\% is a strong result for a first-ever ASR system for
a low-resource language with approximately 14 hours of training audio. For reference,
state-of-the-art WERs for well-resourced languages like English typically fall below 5\% on
standard benchmarks with thousands of hours of data. More relevantly, systems for
comparably low-resource African languages such as Wolof and Swahili, trained with similar
methodologies, have reported WERs in the range of 30--88\%~\cite{pindoh2025selfsupervised},
suggesting that our result is competitive or superior for this category of languages, though
direct comparison is difficult due to dataset differences.
 
Several factors contribute to the residual errors. First, the limited size of the training
corpus (approximately 11\,134 utterances, 13 validated hours) constrains the model's
ability to learn all phonological contrasts of Basa\`{a}. Tonal distinctions in particular
require sufficient examples of minimal pairs to be reliably learned. Second, the dialectal
diversity of Basa\`{a} (twelve dialects) introduces acoustic variation that the model must
generalize across, despite being trained predominantly on one dialect. Third, the
orthographic conventions for Basa\`{a} are not yet as uniform as those for French or
English, introducing inconsistency in the training labels.
 
The notably low CER (3.51\%) relative to WER (14.13\%) suggests that errors are systematic
and localized, often affecting specific characters (diacritics, tonal markers) rather than
producing unintelligible outputs. This pattern indicates that the model has a strong
underlying understanding of Basa\`{a} phonology and primarily struggles with fine-grained
distinctions. Post-processing with a Basa\`{a} character $n$-gram language model or a
tonal correction module could potentially reduce these errors significantly.
 
From a linguistic perspective, the success of the character-level CTC approach validates
the design decision to bypass word-level modeling in favor of character-level decoding.
Given Basa\`{a}'s morphological richness and the absence of a machine-readable Basa\`{a}
lexicon, character-level CTC provides a flexible and linguistically principled framework
that does not presuppose knowledge of the word inventory.

\subsection{Future Directions}
 
Based on the results and observations from this work, we identify several high-priority
directions for future research:
 
\begin{itemize}
  \item \textbf{Corpus expansion:} The most impactful improvement would come from enlarging
        the training corpus, particularly through the Mahop application's data collection
        mechanism. Increasing the corpus to 50+ hours with diverse speakers, dialects, and
        acoustic conditions would likely reduce the WER substantially.
 
  \item \textbf{Tonal language modeling:} Developing a Basa\`{a} $n$-gram or neural
        language model trained on existing Basa\`{a} text (Biblical translations, community
        web content) and integrating it as a post-processor with the acoustic model could
        correct tonally induced errors.
 
  \item \textbf{Dialect adaptation:} Training dialect-specific sub-models or incorporating
        dialect identification into the pipeline would improve robustness across Basa\`{a}'s
        twelve documented dialects.
 
  \item \textbf{Extended training:} Training for additional epochs with learning rate
        scheduling and regularization techniques (weight decay, dropout) could further
        reduce the WER.
 
  \item \textbf{Cloud deployment:} Hosting the Mahop API on a cloud GPU instance would make
        the ASR system publicly accessible and accelerate community-driven data collection.
 
  \item \textbf{Extension to other Cameroonian languages:} The methodology developed here
        is directly applicable to other low-resource Cameroonian languages such as Ewondo,
        Fulfulde, and Ghomala, contributing to the broader digital inclusion of Cameroon's
        linguistic heritage.
\end{itemize}
 
\section{Conclusion}
\label{sec:conclusion}
 
This paper has presented the first ASR system for the Basa\`{a} language, a low-resource
Bantu language spoken by approximately 300\,000 people in Cameroon. We motivated the
system design through a detailed linguistic analysis of Basa\`{a}'s phonological,
morphological, and syntactic properties including its tonal system, vowel quantity
distinctions, rich consonant inventory, and agglutinative morphology and showed how these
properties informed our modeling decisions.
 
Our proposed system fine-tunes the XLS-R-300M model, a large-scale multilingual
self-supervised speech model, on a Basa\`{a}--French corpus from Mozilla Common Voice using
a character-level CTC objective. The system achieves a WER of 14.13\% and a CER of 3.51\%
on a test set of 1\,671 utterances, establishing the first quantitative baseline for
Basa\`{a} ASR. Qualitative analysis confirms that the model achieves perfect transcriptions
on many utterances and that remaining errors are predominantly localized to diacritics,
tonal markers, and word boundaries.

This work demonstrates that modern transfer learning techniques, when paired with
thoughtful linguistic analysis and an appropriate architecture, can yield meaningful ASR
performance even for very low-resource African languages. We hope that it will serve as
both a technical baseline and a methodological blueprint for future ASR research on
Basa\`{a} and on the many other underrepresented languages of Cameroon and Africa.

\bibliographystyle{IEEEtran}
\bibliography{mybibfile.bib}

@article{alabi2025charting,
  author    = {Alabi, Jesujoba O. and Hedderich, Michael A. and Adelani, David Ifeoluwa and Klakow, Dietrich},
  title     = {Charting the Landscape of {African} {NLP}: Mapping Progress and Shaping the Road Ahead},
  journal   = {arXiv preprint arXiv:2505.21315},
  year      = {2025},
  doi       = {10.48550/arXiv.2505.21315}
}

@article{babu2021xlsr,
  author    = {Babu, Arun and Wang, Changhan and Tjandra, Andros and Lakhotia, Kushal and
               Xu, Qiantong and Goyal, Naman and Singh, Kritika and von Platen, Patrick and
               Saraf, Yatharth and Pino, Juan and Baevski, Alexei and Conneau, Alexis and Auli, Michael},
  title     = {{XLS-R}: Self-supervised Cross-lingual Speech Representation Learning at Scale},
  journal   = {arXiv preprint arXiv:2111.09296},
  year      = {2021},
  doi       = {10.48550/arXiv.2111.09296}
}

@article{nzali2022basaa,
  author    = {Nzali, Martial De Tailleur and others},
  title     = {Tone Prediction and Orthographic Conversion for {B}asa{\`a}},
  journal   = {arXiv preprint arXiv:2210.06986},
  year      = {2022},
  doi       = {10.48550/arXiv.2210.06986}
}

@inproceedings{baevski2020wav2vec,
  author    = {Baevski, Alexei and Zhou, Yuhao and Mohamed, Abdelrahman and Auli, Michael},
  title     = {wav2vec 2.0: {A} Framework for Self-Supervised Learning of Speech Representations},
  booktitle = {Advances in Neural Information Processing Systems (NeurIPS)},
  volume    = {33},
  pages     = {12449--12460},
  year      = {2020}
}

@article{besacier2014asr,
  author    = {Besacier, Laurent and Barnard, Etienne and Karpov, Alexey and Schultz, Tanja},
  title     = {Automatic speech recognition for under-resourced languages: {A} survey},
  journal   = {Speech Communication},
  volume    = {56},
  pages     = {85--100},
  year      = {2014},
  doi       = {10.1016/j.specom.2013.07.008}
}

@article{conneau2020unsupervised,
  author    = {Conneau, Alexis and Baevski, Alexei and Collobert, Ronan and Mohamed, Abdelrahman and Auli, Michael},
  title     = {Unsupervised Cross-lingual Representation Learning for Speech Recognition},
  journal   = {arXiv preprint arXiv:2006.13979},
  year      = {2020},
  doi       = {10.48550/arXiv.2006.13979}
}

@article{kamdem2025decolonizing,
  author    = {Kamdem, Sylvain and Ojongnkpot, Christelle and Van Pinxteren, Bart},
  title     = {Decolonizing {Cameroon}'s language policies: {A} critical assessment},
  journal   = {Applied Linguistics Review},
  volume    = {16},
  number    = {2},
  pages     = {1007--1029},
  year      = {2025},
  doi       = {10.1515/applirev-2023-0273}
}

@article{makasso2008intonation,
  author    = {Makasso, Emmanuel-Moselly},
  title     = {Intonation et m{\'e}lismes dans le discours oral spontan{\'e} en {Basa{\`a}}},
  journal   = {TIPA. Travaux interdisciplinaires sur la parole et le langage},
  number    = {27},
  year      = {2008},
  doi       = {10.4000/tipa.394}
}

@article{makasso2015basaa,
  author    = {Makasso, Emmanuel-Moselly and Lee, Seunghun J.},
  title     = {{Basa{\'{a}}}},
  journal   = {Journal of the International Phonetic Association},
  volume    = {45},
  number    = {1},
  pages     = {71--79},
  year      = {2015},
  doi       = {10.1017/S0025100314000383}
}

@article{pindoh2025selfsupervised,
  author    = {Pindoh, Pride Doriane and Yonta, Paulin Melatagia},
  title     = {Self-supervised and multilingual learning applied to the {Wolof}, {Swahili} and {Fongbe}},
  journal   = {Revue Africaine de Recherche en Informatique et Math{\'e}matiques Appliqu{\'e}es},
  volume    = {42},
  year      = {2025},
  url       = {https://inria.hal.science/hal-04547298/}
}

@inproceedings{vaswani2017attention,
  author    = {Vaswani, Ashish and Shazeer, Noam and Parmar, Niki and Uszkoreit, Jakob and
               Jones, Llion and Gomez, Aidan N. and Kaiser, {\L}ukasz and Polosukhin, Illia},
  title     = {Attention is All You Need},
  booktitle = {Advances in Neural Information Processing Systems (NeurIPS)},
  volume    = {30},
  year      = {2017}
}

@article{zhao2022improving,
  author    = {Zhao, Jiatong and Zhang, Wei-Qiang},
  title     = {Improving Automatic Speech Recognition Performance for Low-Resource Languages
               with Self-Supervised Models},
  journal   = {IEEE Journal of Selected Topics in Signal Processing},
  volume    = {16},
  number    = {6},
  pages     = {1227--1241},
  year      = {2022}
}

@techreport{zribi1989preliminaires,
  author      = {Zribi-Hertz, Anne and Ndiki-Mayi, Rolande},
  title       = {Pr{\'e}liminaires {\`a} l'{\'e}tude syntaxique du {Basa{\`a}} ({Cameroun})},
  institution = {Langues et Grammaire},
  number      = {3},
  year        = {1989}
}

@book{champaud2021mom,
  title={Mom: terroir bassa (Cameroun)},
  author={Champaud, Jacques},
  volume={9},
  year={2021},
  publisher={Walter de Gruyter GmbH \& Co KG}
}

 \section*{Appendix}
 
Beyond the scientific contribution, we developed a prototype mobile application named
\textit{Mahop}, integrating the fine-tuned XLS-R model into a real-world speech
transcription interface. The application is built using Flutter (frontend) and Flask
(backend API), with the ASR model served via a local Python server.
 
Mahop supports three user roles: (1)~standard users, who can record speech and receive
real-time Basa\`{a} transcriptions; (2)~contributors, who can record new audio and propose
corresponding transcriptions, thereby expanding the dataset; and (3)~validators, who can
review contributor submissions, accept or reject them, and export validated data. This
crowdsourcing mechanism directly addresses the primary bottleneck of low-resource ASR: the
lack of labeled training data. By enabling community participation in data collection,
Mahop has the potential to grow the Basa\`{a} corpus organically over time.
 
In its current state, the backend is deployed locally, with the Flask server running on a
development machine accessible via IP address. Data are stored in a PostgreSQL database.
Future work will involve deploying the backend on a cloud GPU instance to enable wide-scale
public access. Some screenshots of the application are as follows.


\onecolumn
\begin{itemize}
    \item \textbf{Login/Registration Screen}: In our system, three categories of users can be distinguished, namely simple users, contributors, and annotators.
    
    \begin{figure}[H]
        \centering
        \caption{Registration and login interface}
\vspace{0.5cm}
        \begin{minipage}{0.32\textwidth}
            \includegraphics[width=\linewidth]{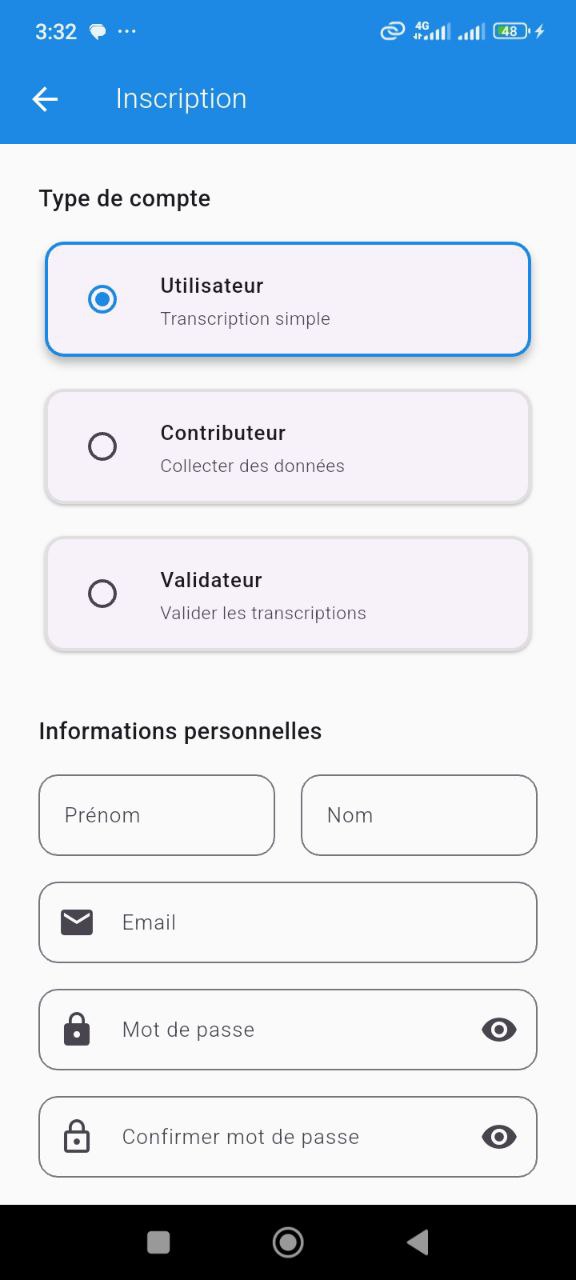}
        \end{minipage}
        \hfill
        \begin{minipage}{0.32\textwidth}
            \includegraphics[width=\linewidth]{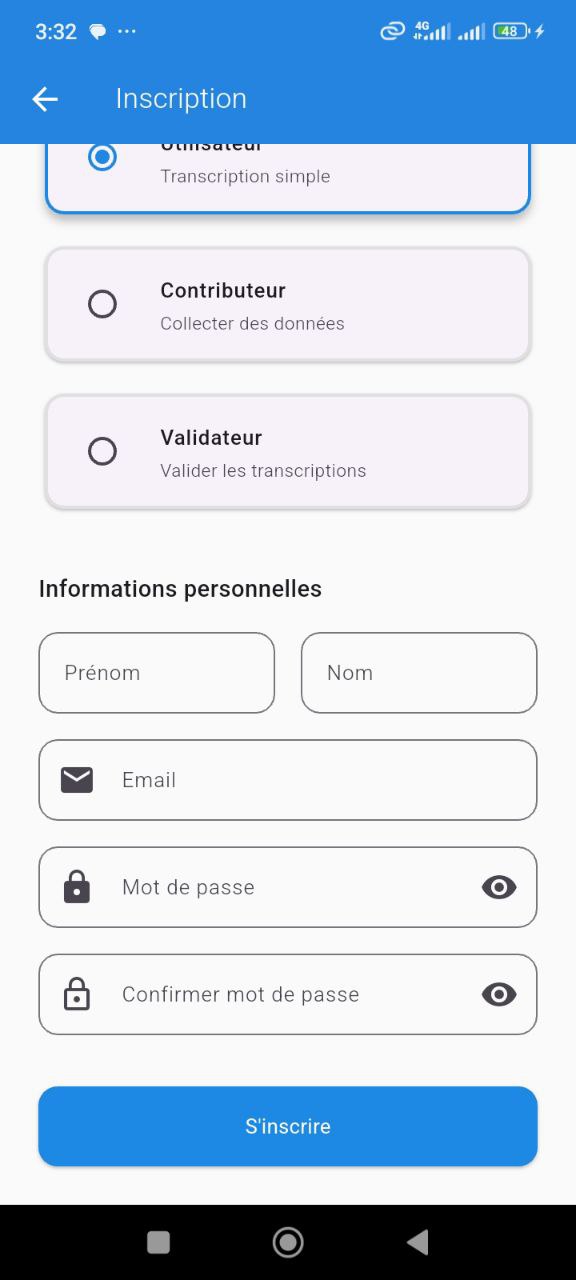}
        \end{minipage}
        \hfill
        \begin{minipage}{0.32\textwidth}
            \includegraphics[width=\linewidth]{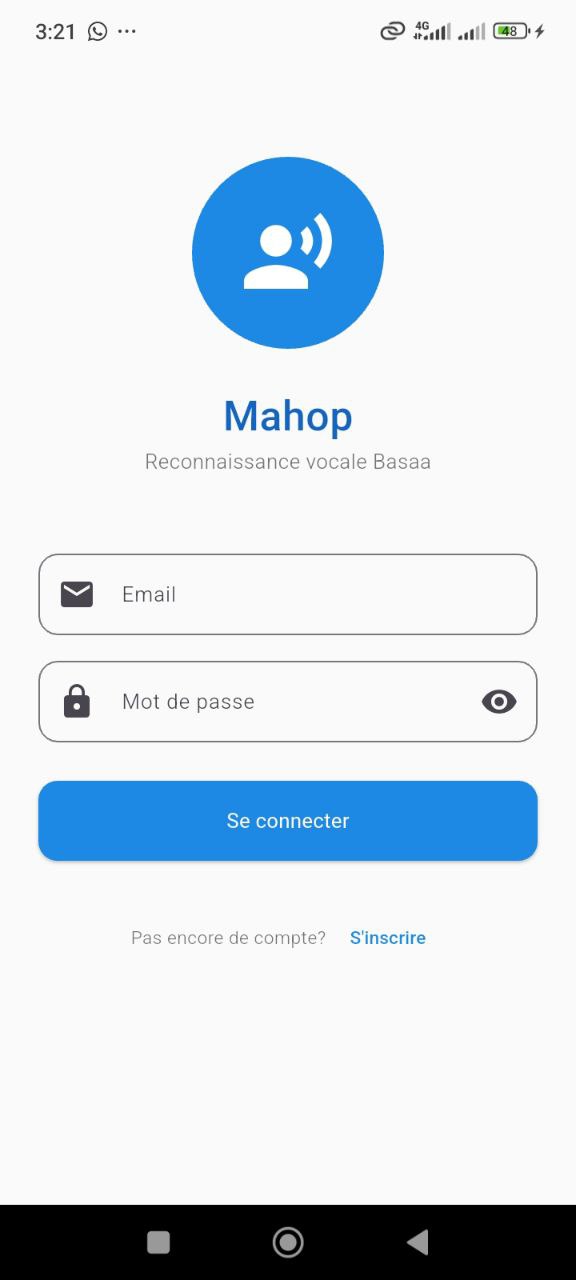}
        \end{minipage}
    \end{figure}
\end{itemize}


\begin{itemize}
    \item \textbf{Simple user view}: the user can perform transcriptions and manage their transcription history.
    
    \begin{figure}[H]
        \centering
  \caption{Simple user interface}
\vspace{0.5cm}
        \begin{minipage}{0.32\textwidth}
            \includegraphics[width=\linewidth]{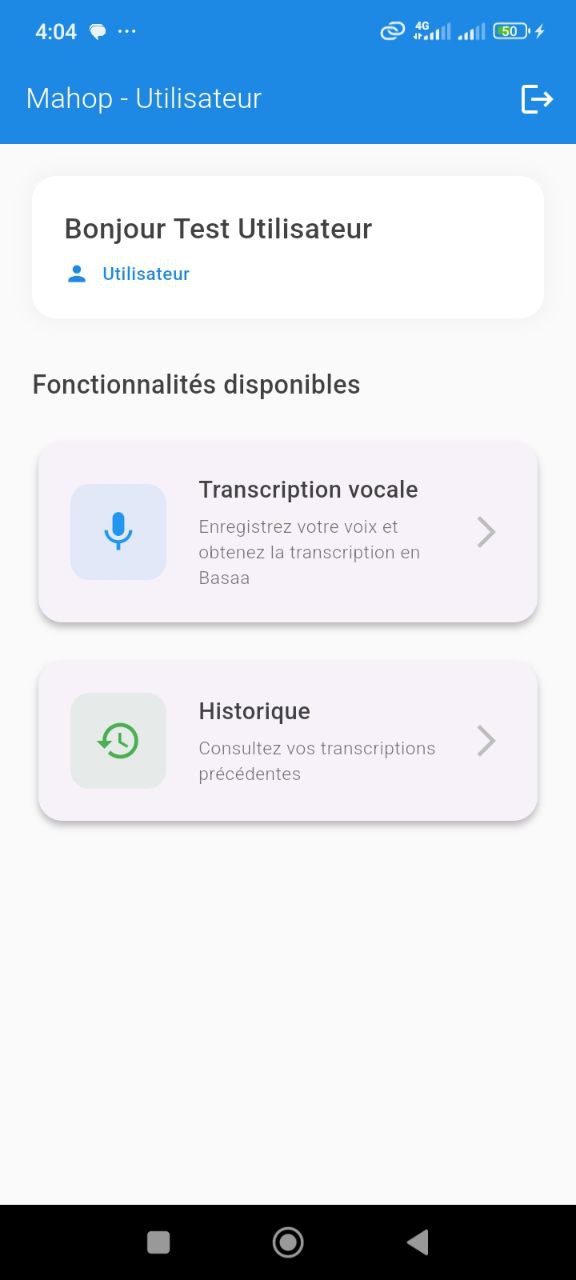}
        \end{minipage}
        \hfill
        \begin{minipage}{0.32\textwidth}
            \includegraphics[width=\linewidth]{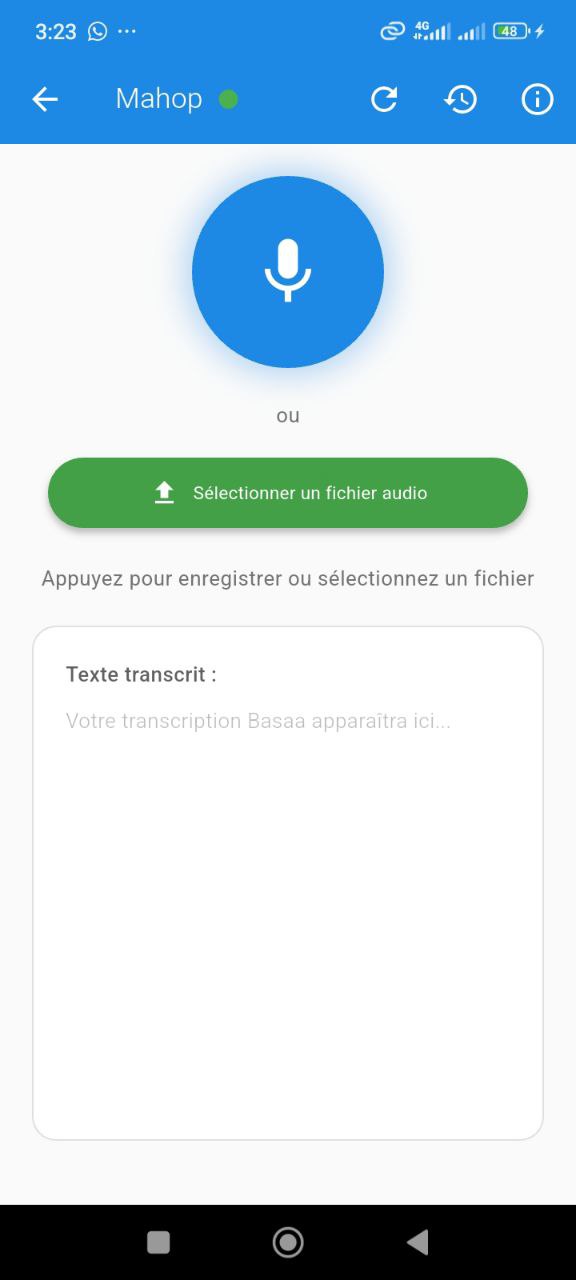}
        \end{minipage}
        \hfill
        \begin{minipage}{0.32\textwidth}
            \includegraphics[width=\linewidth]{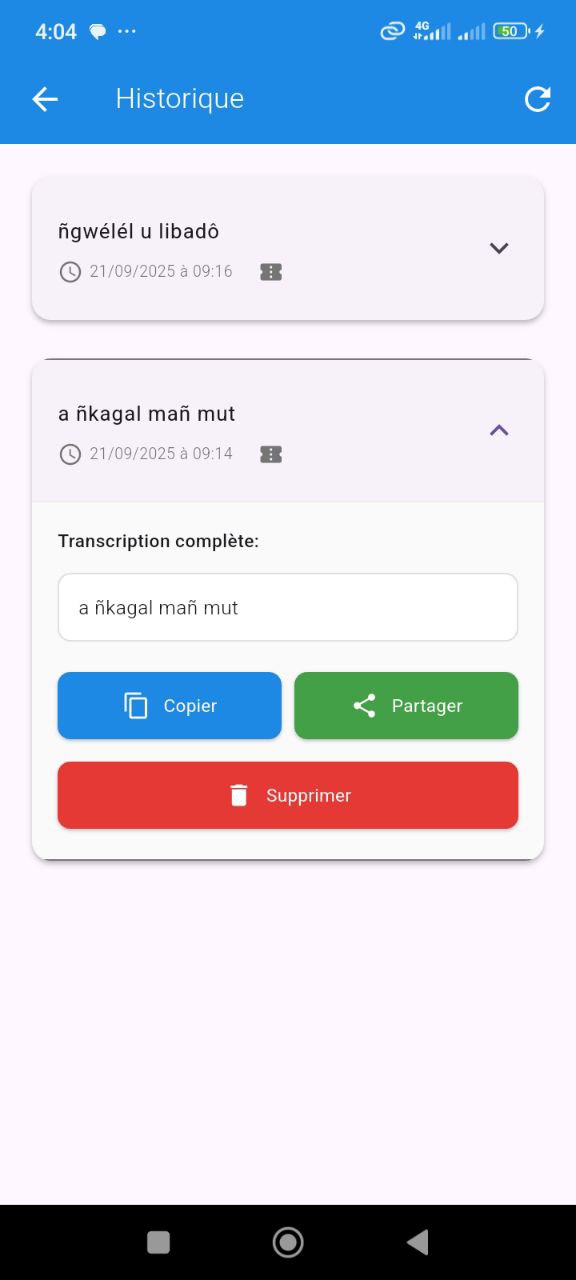}
        \end{minipage}
    \end{figure}
\end{itemize}


\begin{itemize}
    \item \textbf{Contributor view}: the contributor can add a new contribution and view their different contributions.
    
    \begin{figure}[H]
        \centering
        \caption{Contributor Interface}
\vspace{0.5cm}
        \begin{minipage}{0.32\textwidth}
            \includegraphics[width=\linewidth]{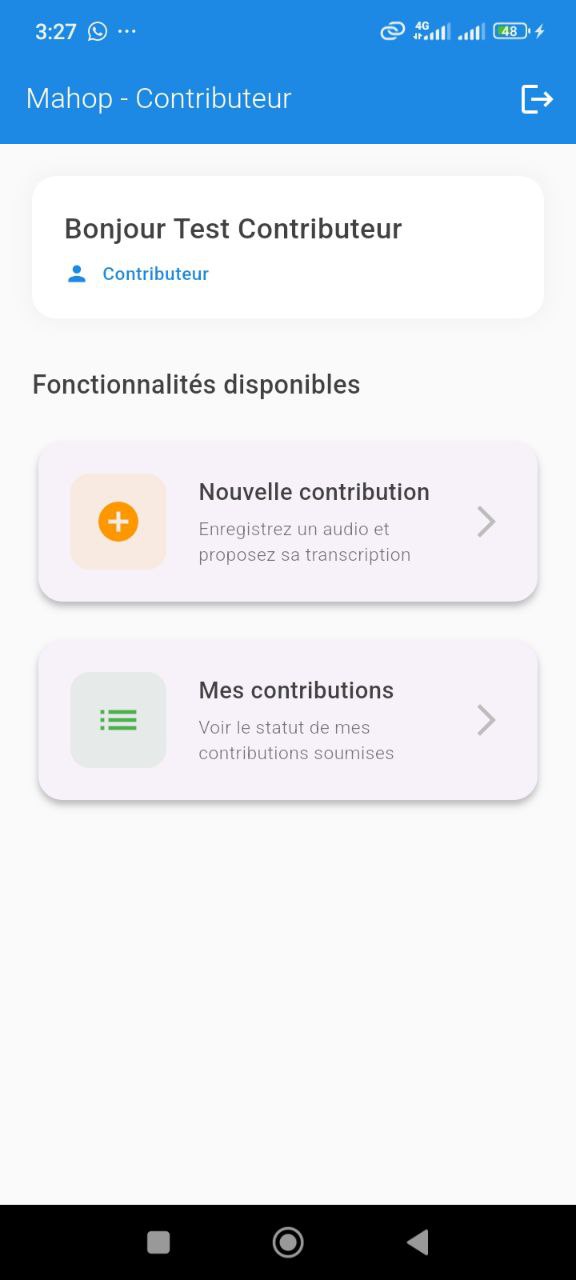}
        \end{minipage}
        \hfill
        \begin{minipage}{0.32\textwidth}
            \includegraphics[width=\linewidth]{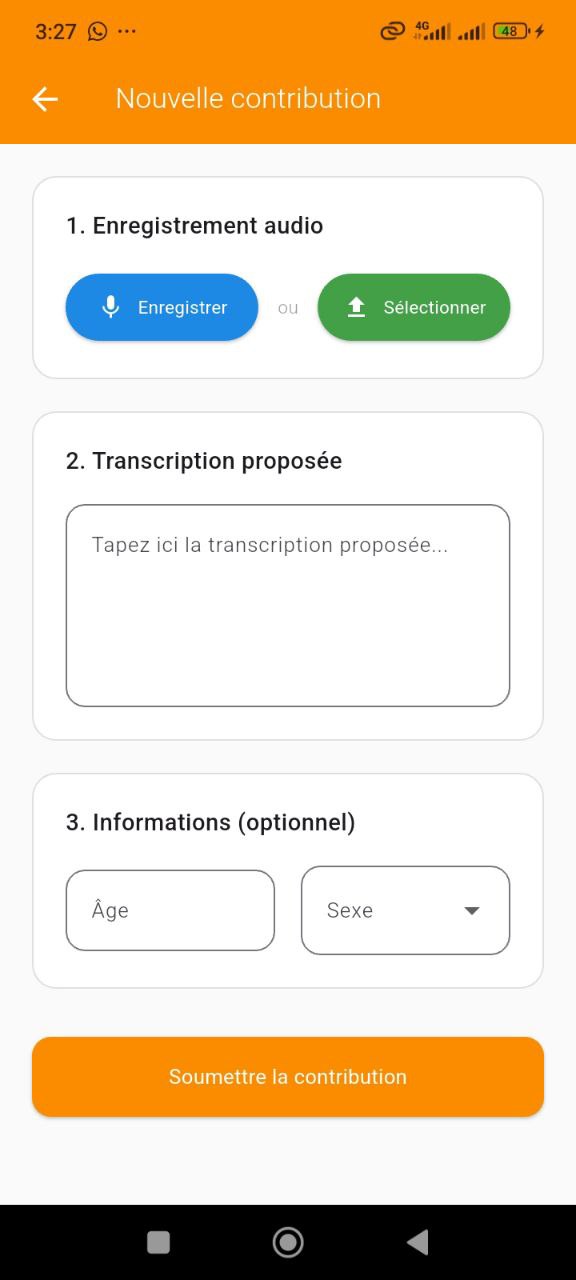}
        \end{minipage}
        \hfill
        \begin{minipage}{0.32\textwidth}
            \includegraphics[width=\linewidth]{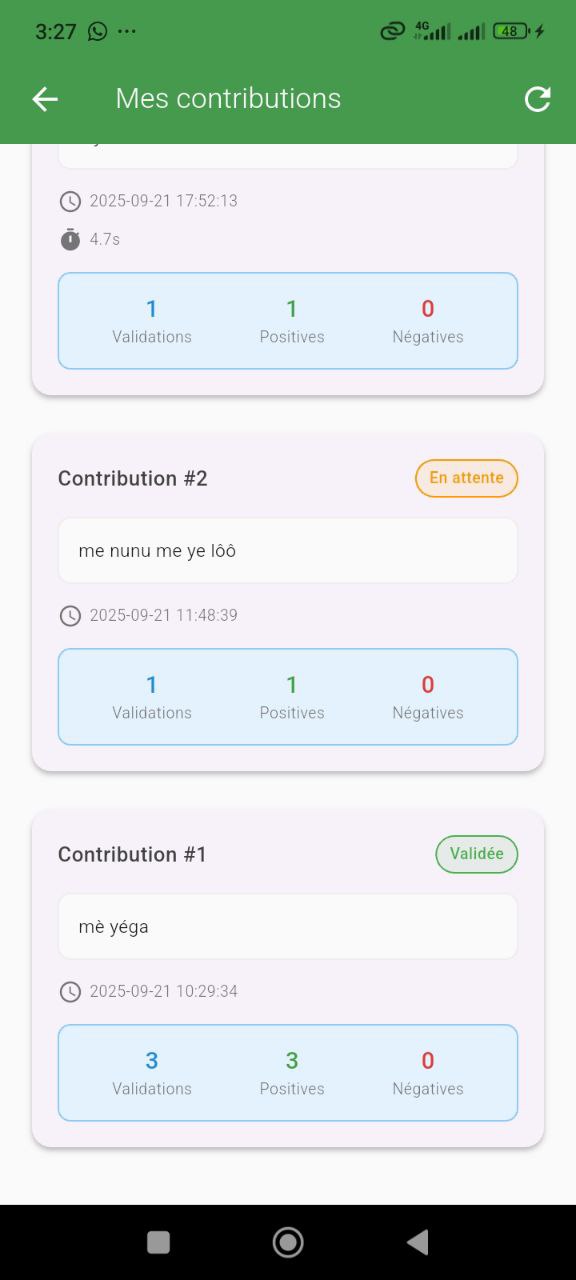}
        \end{minipage}
    \end{figure}
\end{itemize}


\begin{itemize}
    \item \textbf{Validator view}: The validator can review and validate contributions, view validation statistics, and export validated data in Excel format.
    
    \begin{figure}[H]
        \centering
        \caption{Validator Interface}
\vspace{0.5cm}
        \begin{minipage}{0.32\textwidth}
            \includegraphics[width=\linewidth]{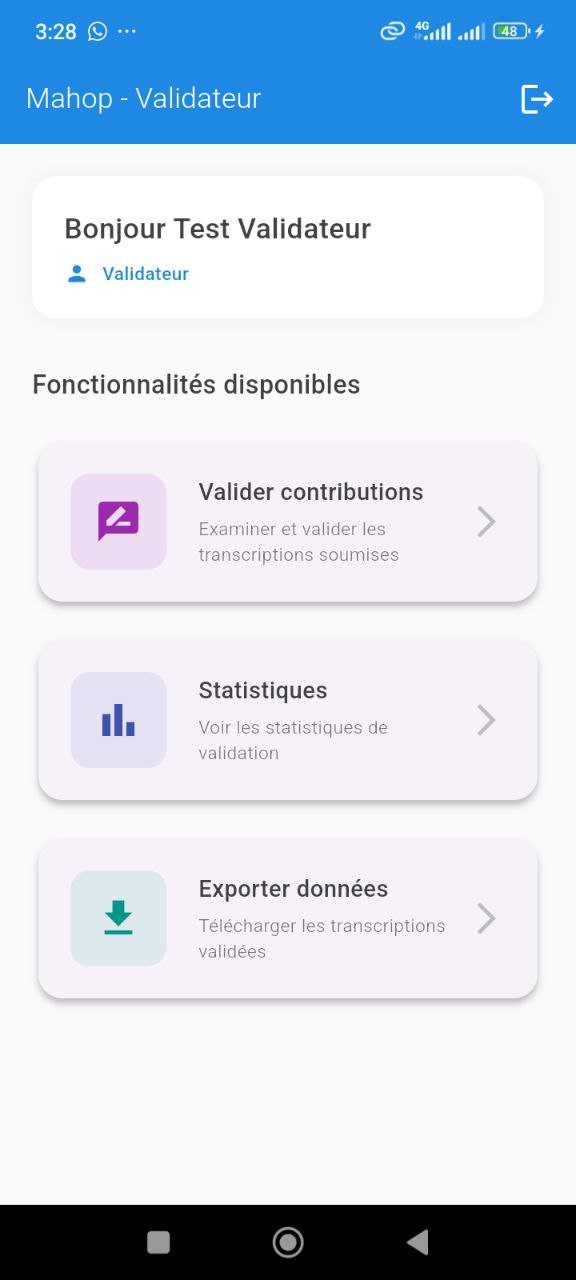}
        \end{minipage}
        \hfill
        \begin{minipage}{0.32\textwidth}
            \includegraphics[width=\linewidth]{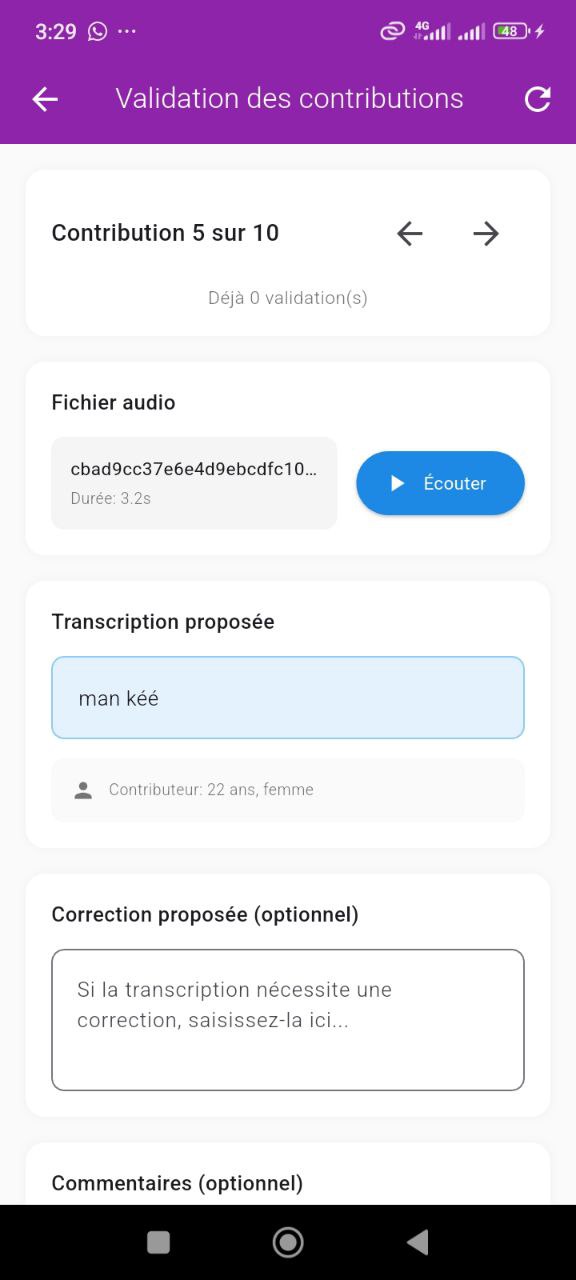}
        \end{minipage}
        \hfill
        \begin{minipage}{0.32\textwidth}
            \includegraphics[width=\linewidth]{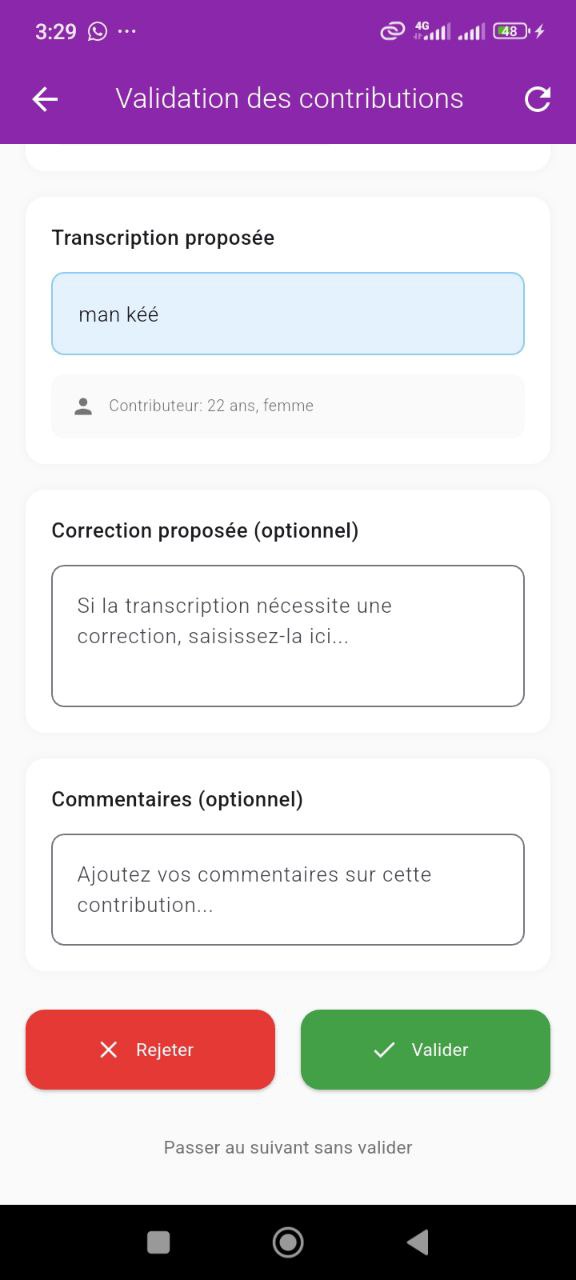}
        \end{minipage}
    \end{figure}


    \begin{figure}[H]
        \centering
        \caption{Validator interface}
\vspace{0.5cm}
        \begin{minipage}{0.32\textwidth}
            \includegraphics[width=\linewidth]{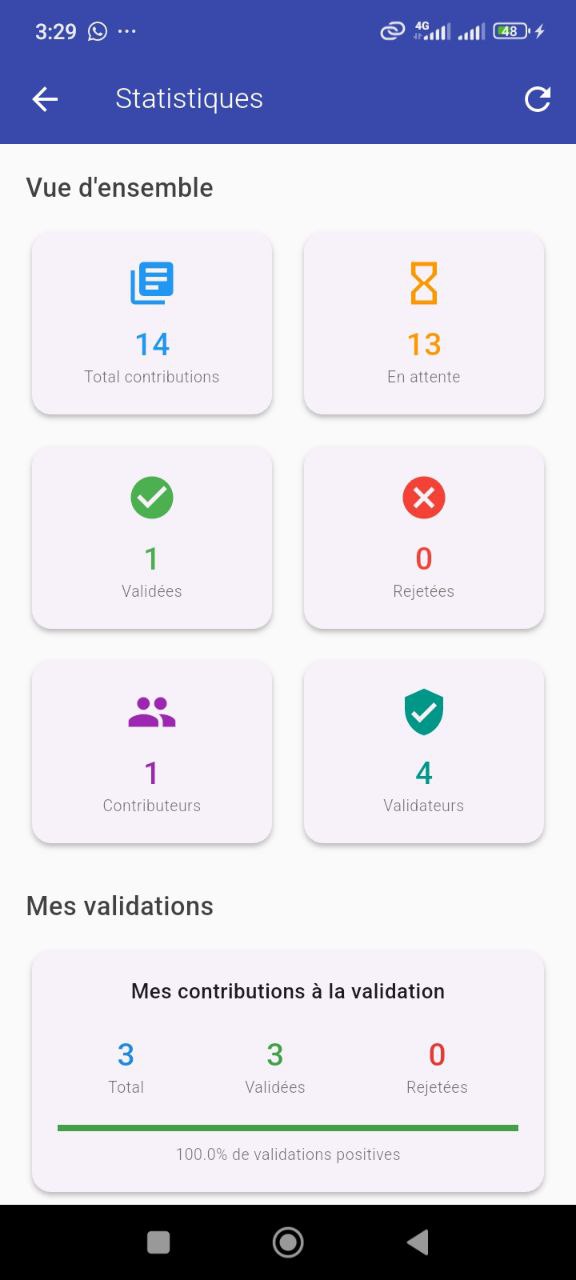}
        \end{minipage}
        \hfill
        \begin{minipage}{0.32\textwidth}
            \includegraphics[width=\linewidth]{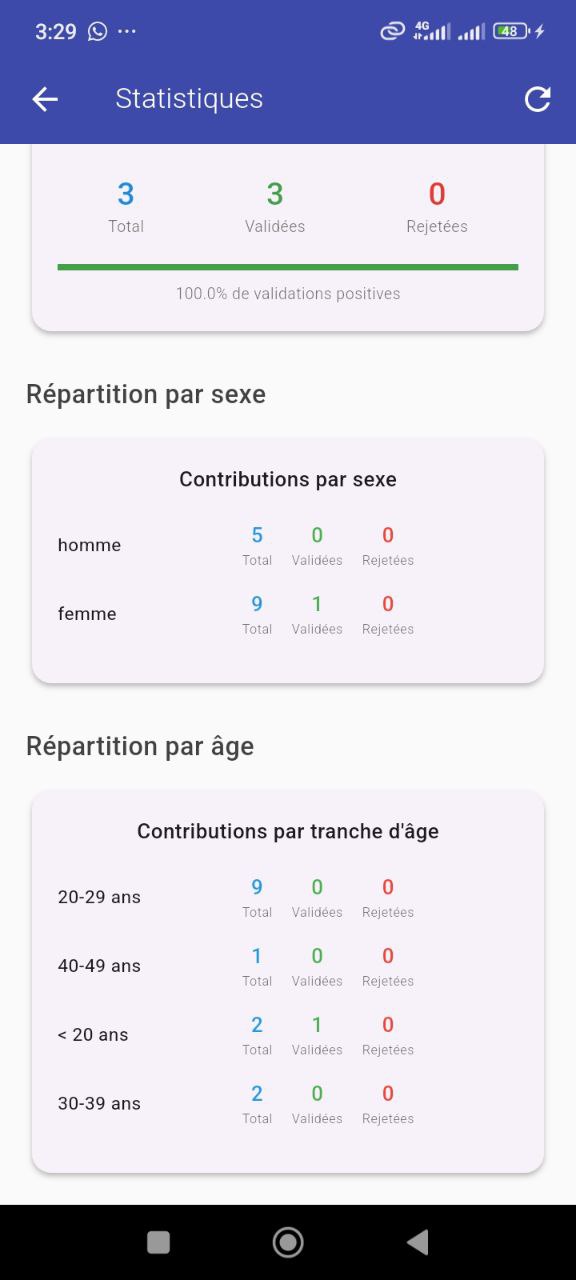}
        \end{minipage}
        \hfill
        \begin{minipage}{0.32\textwidth}
            \includegraphics[width=\linewidth]{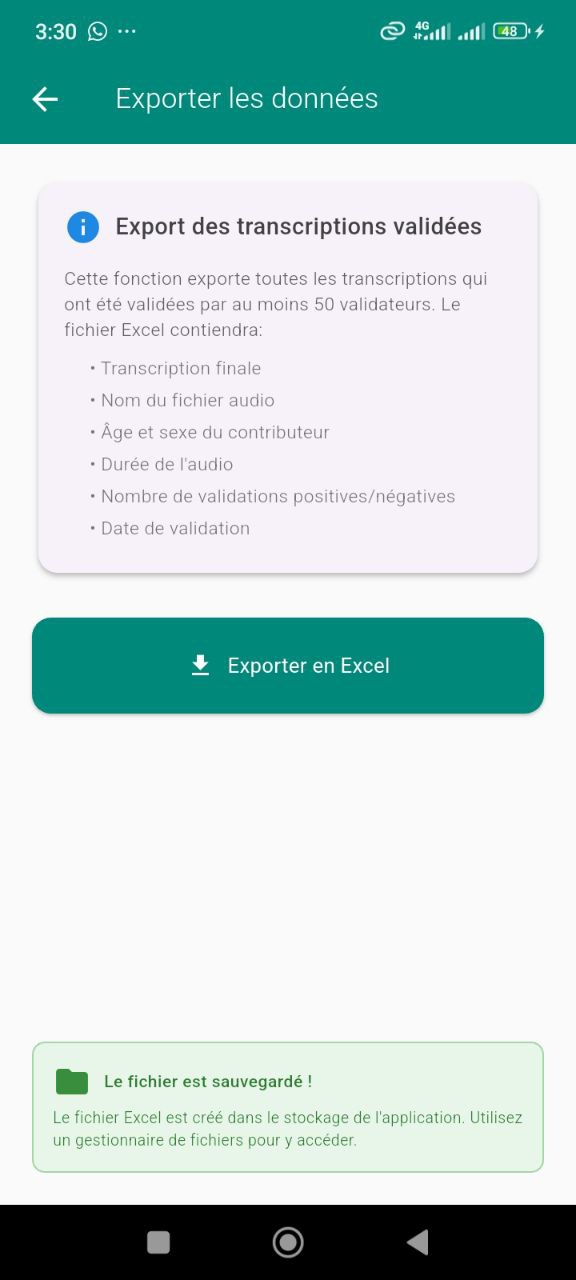}
        \end{minipage}
    \end{figure}
\end{itemize}
\EOD
\end{document}